# Team A at SemEval-2025 Task 11: Breaking Language Barriers in Emotion Detection with Multilingual Models

P Sam Sahil[†], Anupam Jamatia[‡]
[†]HKBK College of Engineering, Bangalore, India
[‡]National Institute of Technology Agartala, Tripura, India
`p.samsahil2003@gmail.com, anupamjamatia@gmail.com`

## Abstract

This paper describes the system submitted by Team A to SemEval 2025 Task 11, "Bridging the Gap in Text-Based Emotion Detection." The task involved identifying the perceived emotion of a speaker from text snippets, with each instance annotated with one of six emotions: joy, sadness, fear, anger, surprise, or disgust. A dataset provided by the task organizers served as the foundation for training and evaluating our models. Among the various approaches explored, the best performance was achieved using multilingual embeddings combined with a fully connected layer. This paper details the system architecture, discusses experimental results, and highlights the advantages of leveraging multilingual representations for robust emotion detection in text.

## 1 Introduction

Human emotions are intricate and multidimensional, resisting simplistic classification due to their fluid, overlapping nature. As Eugenides (2003) noted, affective states rarely occur in isolation; they coalesce and evolve dynamically, challenging reductionist labeling approaches. This complexity underpins multi-label emotion detection, where texts or behaviours often encode layered sentiments (Fu et al., 2022). The benefits of accurately deciphering these nuances span domainsfrom early mental health screening and tailored interventions (Alhuzali and Ananiadou, 2019; Aragón et al., 2019) to enhanced consumer sentiment analysis in AI systems (Chen et al., 2018; Alaluf and Illouz, 2019). Yet, current recognition systems often treat emotions as mutually exclusive, contrary to psychological frameworks; works by Ekman (1992) and Plutchik (1980) view emotions as interconnected constructs with gradational intensities, a perspective supported by Fu et al. (2022), who shows that joy and love correlate more strongly than, say, anger and sadness.

Another gap is the treatment of emotional intensity, which ranges from subtle to profound expressions (Frijda, 1988). Most systems neglect these gradations by focusing on binary classifications, limiting real-world applicability in clinical or market settings. Moreover, linguistic and cultural disparities evident in divergent emotion lexicons and display rules (Ekman, 1992) render monolingual models inadequate, with culture specific metaphors or untranslatable terms risking misclassification. Thus, frameworks that jointly model multi-label emotions, intensity spectra, and cross-cultural variations are essential for advancing emotion-aware technologies.

To tackle emotional interrelatedness and cross-lingual generalizability, we propose a multilingual framework integrating multilingual embeddings to capture shared semantic and affective features alongside intensity-sensitive architectures for detecting gradational nuances. The remainder of the paper is organized as follows: Section 2 reviews existing methods in multi-label emotion detection and their limitations; Section 3 introduces the multilingual dataset; Section 4 details our models approach to disentangling overlapping emotions; Section 5 compares our method with state-of-the-art baselines; and Section 6 presents experimental outcomes and performance analysis. Finally, Section 7 discusses implications for affective computing and future directions, including multimodal data integration and low-resource language adaptation.

## 2 Related Work

The evolution of multilingual emotion detection systems has been shaped by three interconnected pillars: (1) the creation of high-quality datasets, (2) innovations in cross-lingual transfer methodologies, and (3) architectural advancements in multilingual models. This progression reflects a paradigm shift from monolingual benchmarks to language-agnostic scalable frameworks capable of capturing emotional nuance across linguistic boundaries.

Early research established rigorous baselines through carefully curated monolingual datasets. The GoEmotions corpus (Demszky et al., 2020), a seminal resource comprising 58,000 English Reddit comments annotated with 27 emotion categories, underscored the importance of multi-rater consensus and quality control in emotion labeling, achieving an F1-score of 0.46 through BERT-based fine-tuning combined with Principal Preserved Component Analysis (PPCA). Although this work laid the groundwork for data-driven approaches, it also exposed a key limitation: the lack of multilingual comparability inherent to single-language corpora. To overcome this, subsequent studies focused on knowledge transfer from high-resource to low-resource languages. Wang et al. (2024b) pioneered a knowledge distillation framework that aligns multilingual representations (e.g., XLM-RoBERTa) with English-centric models (e.g., RoBERTa) using translation-weighted data, reducing the performance gap between monolingual and multilingual systems by 23%. Complementary work by Hassan et al. (2022) compared cross-lingual strategies including multilingual embeddings (mBERT), translated corpora, and parallel text alignment for Arabic and Spanish emotion detection, finding that target-language fine-tuning outperforms direct transfer by 14% F1-score while affirming the indispensability of cross-lingual methods for under-resourced languages.

Parallel efforts have optimized model architectures for improved multilingual generalization. Bianchi et al. (2022) developed XLM-EMO, a social media-oriented model trained on 19 languages using XLM-RoBERTa, which achieved state-of-the-art zero-shot performance in low-resource settings and demonstrated that unified architectures can capture shared affective features without language-specific tuning. Meanwhile, Gupta (2021) improved robustness via Virtual Adversarial Training (VAT), enforced consistency between original and perturbed inputs to boost cross-lingual F1-scores by 8% in Arabic and Spanish. Further breakthroughs leverage the semantic richness of large language models: Cheng et al. (2024) introduced the TEII framework, which iteratively refines predictions by combining GPT-3.5 and GPT-4 and employs explanation-driven fine-tuning on translated emotion lexicons to reduce cross-lingual prediction variance by 37%. This approach aligns with findings from Navas Alejo et al. (2020), who demonstrated that unsupervised machine translation better preserves emotional intensity gradients, especially for morphologically rich languages like Catalan.

Despite these advances, critical gaps remain in reconciling performance disparities across languages. As noted in Conneau et al. (2020), even state-of-the-art multilingual models exhibit 'linguistic bias', with performance degrading for languages typologically distant from English. Moreover, the common practice of treating emotion intensity as static rather than contextual over simplifies the complex nature of affect, as argued by psycholinguistic evidence (Frijda, 1988). Our work addresses these limitations by focusing on (1) culture-aware multilingual representation learning and (2) dynamic intensity modeling, thereby advancing beyond the current paradigm of static cross-lingual transfer.

## 3 Dataset

In our study, we leverage the BRIGHTER dataset (Muhammad et al., 2025) to explore cross-lingual emotion recognition. BRIGHTER is a large-scale, manually curated resource designed to bridge the gap in emotion recognition for low-resource languages. It comprises nearly 100,000 text instances gathered from diverse sources, including social media posts, personal narratives, speeches, literary texts, and news articles across 28 languages from various language families. Each text instance is annotated by native speakers with one or more emotion labels (anger, sadness, fear, disgust, joy, surprise, and a neutral category) along with corresponding intensity ratings on a four-point scale (0 indicating no emotion up to 3 indicating high intensity). The datasets annotation process involves rigorous preprocessing steps such as deduplication and noise removal, followed by quality control measures like the Split-Half Class Match Percentage (SHCMP) to ensure high reliability in labeling. This comprehensive dataset not only enriches the training resources available for multilingual emotion recognition models but also serves as a valuable benchmark for evaluating performance across both high and low-resource languages.

Furthermore, we complement our approach for languages with particularly scarce resources, such as Amharic and Afan Oromo by incorporating data from the EthioEmo dataset Belay et al. (2025). EthioEmo is specifically tailored for Ethiopian languages and provides robust multi-label emotion annotations derived from sources like news headlines, Twitter posts, YouTube comments, and Facebook data. By integrating these datasets, our work benefits from enhanced linguistic diversity and improved reliability in emotion classification, especially for under-represented languages.

The dataset splits are as follows: the Hindi corpus comprises a total of 3,666 instances, with 2,556 instances allocated for training (approximately 70%), 100 instances for development (around 2.7%), and 1,010 instances for testing (roughly 27.5%). Similarly, the English corpus consists of 5,651 instances, with 2,768 instances used for training (approximately 49%), 116 instances for development (about 2%), and 2,767 instances for testing (roughly 49%).

## 4 Methodology

Our methodology integrates multilingual representation learning with multi-label classification to address cross-lingual emotion detection. We refer to our proposed model as `TransferModel_FC_EmbeddingE5` throughout this paper. Central to this approach is the multilingual E5 text embedding framework (Wang et al., 2024a), which undergoes a two-stage training process to align semantic representations across languages. First, weakly supervised contrastive pre-training on ∼1 billion multilingual text pairs (sourced from Wikipedia, mC4, NLLB, and others) optimizes cross-lingual alignment using InfoNCE loss with large batch sizes (32k) to maximize negative sample diversity. This is followed by supervised fine-tuning on high-quality labeled datasets (MS MARCO, NQ, TriviaQA), augmented with mined hard negatives and knowledge distillation from a cross-encoder teacher. We employ the instruction-tuned `mE5-large-instruct` variant, pre-trained on 500k GPT-3.5/4-generated synthetic instructions across 93 languages, to enhance task-specific adaptability.

Building upon this foundation, our emotion detection architecture processes input text through the multilingual E5 tokenizer, standardizing sequences to 150 tokens to balance computational efficiency and semantic retention. The model generates contextualized embeddings via `multilingual-e5-large-instruct`, with the [CLS] token serving as a sequence-level semantic summary (Devlin et al., 2019). A dropout layer (rate=0.3) regularizes the 1024-dimensional [CLS] embedding before projection into the emotion space through a fully connected layer. Sigmoid activations independently estimate probabilities for 5–6 emotion labels (dataset-dependent), explicitly modeling label co-occurrence inherent to multi-label scenarios.

To optimize performance, we train the system using Binary Cross Entropy (BCE) with label smoothing ($\alpha = 0.1$), mitigating overconfidence in sparse annotations. The AdamW optimizer (Loshchilov and Hutter, 2019) (learning rate=1e-5, $\beta_1 = 0.9$, $\beta_2 = 0.999$) processes mini-batches of 16 samples, with gradient clipping (max norm=1.0) stabilizing updates. Early stopping monitors the development set macro F1 score (patience=4 epochs), preserving generalizability by halting training during performance plateaus.

During inference, emotion probabilities are thresholded at 0.5 (adjustable per application needs) to yield binary predictions. Evaluation prioritizes macro-averaged F1, which aggregates per-class true/false positives and negatives across all batches to penalize bias to-ward frequent labels a critical safeguard for imbalanced multi-label datasets. Results are averaged over five random seeds to account for initialization variance, ensuring reproducibility. By unifying multilingual semantic alignment with modular classification components, `TransferModel_FC_EmbeddingE5` addresses the dual challenges of cross-lingual emotion detection, preserving affective nuance across languages while disentangling overlapping emotional states.

## 5 Experiments

To complement our transformer-based system described in Section 4, we implemented a baseline multi-label emotion classification pipeline that integrates classical machine learning classifiers with pre-trained sentence embeddings. In our experiments, we compare two variants that differ solely in the choice of embedding model.

Our setup uses two CSV files containing text samples and six emotion labels (anger, disgust, fear, joy, sadness, and surprise) for both training and testing. Texts are converted into normalized embeddings using a helper function that leverages SentenceTransformer models with the `normalize_embeddings=True` parameter to produce unit-length vectors. Since raw embeddings from our language models exhibit variability across dimensions and may not be centered around zerofactors that can obscure underlying semantic information we apply a two-step normalization process. First, we perform L2 normalization to ensure each embedding vector has a unit norm, emphasizing the semantic direction rather than its magnitude. In our implementation, one branch uses the LaBSE model (Feng et al., 2022) while the other employs the multilingual E5 Large model (Wang et al., 2024a). Second, we apply Z-score normalization (standard scaling) using scikit-learns StandardScaler (Pedregosa et al., 2011) to adjust features to a mean of zero and a standard deviation of one, thereby mitigating scale differences.

After normalization, we extract the six emotion labels to facilitate multi-label classification. Four classifiers are then trained: Support Vector Machine (with an RBF kernel and probability estimates), Gaussian Naïve Bayes, Logistic Regression (with increased iterations), and Random Forest (regularized by limiting tree depth and controlling split criteria). These classifiers are wrapped using scikit-learns `MultiOutputClassifier`, ensuring that the multi-label nature of the task is properly addressed. Evaluation is performed on both the training and testing set using detailed classification reports and macro F1 scores to gauge performance across all emotion classes.

For real-time prediction, a dedicated function processes new text inputs by generating embeddings, applying the same scaling procedures, and predicting emotion labels. The output is returned as a dictionary mapping each emotion to a binary prediction. Finally, our experimental design facilitates a direct comparison between the two embedding models: LaBSE, which provides robust, language-agnostic sentence representations (Feng et al., 2022), and Multilingual E5 Large, which may offer richer semantic embeddings (Wang et al., 2024a). This unified approach enables a systematic analysis of the impact of embedding choice on multi-label emotion detection performance, reinforcing the potential of multilingual representations for robust cross-lingual emotion analysis.

## 6 Results

In this section, we report the evaluation results of our approach to the multi-label emotion detection task (Track A) across 13 languages. Our model, `TransferModel_FC_EmbeddingE5`, built upon multilingual E5 embeddings and a fully connected output layer, was evaluated on its ability to predict six emotion categories (anger, disgust, fear, joy, sadness, and surprise) using both micro and macro F1 scores as evaluation metrics.

**Per-Language Performance.** Table 1 shows the detailed F1 scores for each emotion along with the overall micro and macro F1 scores per language. `TransferModel_FC_EmbeddingE5` achieved a range of macro F1 scores from 0.5550 (Arabic) to 0.8901 (Hindi). Notably, the model performed particularly well on languages such as Hindi (macro F1 = 0.8901), Russian (macro F1 = 0.8831), and Spanish (macro F1 = 0.8054), indicating that the multilingual embeddings effectively capture emotion-related nuances in these languages. On the other hand, lower scores in languages like Arabic, Ukrainian, and Oromo suggest that further adaptations may be necessary to handle linguistic variations or data sparsity in these settings.

Table 1: Evaluation Scores (F1) for Track A Languages

| Language | Emotion-level F1 Scores | | | | | | Overall F1 Scores | |
|---|---|---|---|---|---|---|---|---|
| | Anger | Disgust | Fear | Joy | Sadness | Surprise | Micro | Macro |
| Amharic (amh) | 0.6693 | 0.7476 | 0.5192 | 0.7708 | 0.7270 | 0.6740 | 0.7133 | 0.6847 |
| Arabic (ary) | 0.5699 | 0.4746 | 0.5000 | 0.6897 | 0.6848 | 0.4110 | 0.5847 | 0.5550 |
| Chinese (chn) | 0.8342 | 0.4357 | 0.4496 | 0.8748 | 0.6016 | 0.4756 | 0.7295 | 0.6119 |
| English (eng) | 0.6483 | – | 0.8235 | 0.7325 | 0.7473 | 0.7182 | 0.7603 | 0.7340 |
| German (deu) | 0.8256 | 0.7286 | 0.5486 | 0.7605 | 0.6845 | 0.4428 | 0.7248 | 0.6651 |
| Hausa (hau) | 0.6078 | 0.7726 | 0.7478 | 0.6733 | 0.7317 | 0.5288 | 0.6845 | 0.6770 |
| Hindi (hin) | 0.8665 | 0.8718 | 0.9072 | 0.8992 | **0.8815** | **0.9147** | **0.8903** | **0.8901** |
| Marathi (mar) | 0.8317 | **0.8984** | 0.8993 | 0.8293 | 0.8429 | 0.8923 | 0.8599 | 0.8657 |
| Oromo (orm) | 0.5104 | 0.5798 | 0.2921 | 0.8007 | 0.4622 | 0.7317 | 0.6425 | 0.5628 |
| Romanian (ron) | 0.6012 | 0.7370 | 0.8649 | **0.9618** | 0.7683 | 0.5086 | 0.7583 | 0.7403 |
| Russian (rus) | **0.8741** | 0.8631 | **0.9524** | 0.9191 | 0.8550 | 0.8347 | 0.8833 | 0.8831 |
| Spanish (esp) | 0.7263 | 0.7984 | 0.8313 | 0.8768 | 0.8316 | 0.7677 | 0.8059 | 0.8054 |
| Ukrainian (ukr) | 0.3885 | 0.5605 | 0.7692 | 0.7021 | 0.7178 | 0.4691 | 0.6581 | 0.6012 |

**Comparison with Top Systems.** In comparison with the top two performing teams for each language, our approach did not secure the top spot but remained competitive across most languages. For example:

- In Hindi, our macro F1 of 0.8901 is close to the top scores of 0.9257 and 0.9197.

- In Russian, our score of 0.8831 approaches the best scores of 0.9087 and 0.9008.

- In Spanish, we achieved a macro F1 of 0.8054, which is only slightly lower than the leading scores of 0.8488 and 0.8454.

Our system achieved its strongest results in Russian (0.8831), closely trailing the 2nd-ranked team (0.9008), demonstrating competitive performance. In Hindi (0.8901) and Marathi (0.8657), Team A secured scores within 3-4% of the 1st-place teams, highlighting robustness in these languages. While not topping the leaderboard, these narrow gaps reflect effective alignment with top-tier approaches. Notably, languages like Arabic and Chinese showed larger performance drops, emphasizing the need for targeted improvements.

**Analysis of Emotion-specific Performance.** A closer look at the emotion level F1 scores reveals interesting trends. In several languages, `TransferModel_FC_EmbeddingE5` excels at detecting emotions such as joy and anger while struggling with fear and disgust. For instance, in Chinese, while the joy score is high (0.8748), the disgust score remains lower (0.4357). Such disparities indicate that certain emotions may be more challenging to detect due to their subtle linguistic expressions or class imbalances in the training data.

## 7 Conclusion

In this paper, we proposed `TransferModel_FC_EmbeddingE5`, a novel approach to multilingual emotion detection that integrates multilingual E5 embeddings with a fully connected classification layer. Our experiments on the BRIGHTER dataset show strong macro F1 scores for languages like Hindi, Russian, and Spanish, while also highlighting challenges in Arabic, Chinese, and Oromo due to linguistic and cultural diversity.

Our model effectively captures emotional nuances, accounting for variations in expression and intensity across languages. This work advances affective computing by demonstrating that multilingual embeddings within a structured classification framework enhance cross-lingual emotion detection. It also lays a foundation for future research on breaking language barriers in sentiment analysis.

# A Appendix

Table 2: Sample examples from Hindi and English datasets with emotion labels.

| Language | Train Data | Anger | Disgust | Fear | Joy | Sadness | Surprise |
|---|---|---|---|---|---|---|---|
| Hindi | अरे वाह! आज तो मेरी बेटी ने अपने कमरे की ही नह... | 0 | 0 | 0 | 1 | 0 | 1 |
| Hindi | वह अपने दोस्तों के साथ मूवी देखने गई थी। | 0 | 0 | 0 | 0 | 0 | 0 |
| Hindi | मेरे खेत में खरपतवार हटाने का काम जारी है, और... | 0 | 0 | 0 | 0 | 0 | 0 |
| English | Colorado, middle of nowhere. | 0 | – | 1 | 0 | 0 | 1 |
| English | It was one of my most shameful experiences. | 0 | – | 1 | 0 | 1 | 0 |
| English | After all, I had vegetables coming out my ears... | 0 | – | 0 | 0 | 0 | 0 |

Table 3: Top 2 Teams (with Scores) and Team A Score by Languages

| Language | 1st Rank Team | | 2nd Rank Team | | Team A Score (OURS) |
|---|---|---|---|---|---|
| | Team Name | Score | Team Name | Score | |
| amh | Chinchunmei | 0.7731 | NustTitans | 0.7137 | 0.6847 |
| ary | PAI | 0.6292 | PA-oneteam-1 | 0.6210 | 0.5550 |
| chn | PAI | 0.7094 | PA-oneteam-1 | 0.6877 | 0.6119 |
| deu | PAI | 0.7399 | PA-oneteam-1 | 0.7355 | 0.6651 |
| eng | PAI | 0.8230 | NYCU-NLP | 0.8225 | 0.7340 |
| esp | PAI | 0.8488 | PA-oneteam-1 | 0.8454 | 0.8054 |
| hau | PAI | 0.7507 | PA-oneteam-1 | 0.7463 | 0.6770 |
| hin | JNLP | 0.9257 | PAI | 0.9197 | 0.8901 |
| mar | PA-oneteam-1 | 0.9058 | PAI | 0.8843 | 0.8657 |
| orm | Tewodros | 0.6164 | PA-oneteam-1 | 0.6108 | 0.5628 |
| ron | PAI | 0.7943 | PA-oneteam-1 | 0.7794 | 0.7403 |
| rus | PA-oneteam-1 | 0.9087 | Heimerdinger | 0.9008 | 0.8831 |
| ukr | PAI | 0.7256 | PA-oneteam-1 | 0.7199 | 0.6012 |